\documentclass[review]{elsarticle}

\usepackage{hyperref}
\usepackage{array}
\usepackage{url}
\usepackage{xcolor}
\usepackage{amsfonts}
\usepackage{multirow}
\usepackage{pdflscape}
\usepackage{afterpage}
\usepackage{capt-of}
\usepackage{textcomp} 
\usepackage{mathtools} 
\usepackage{rotating}

\newcommand\rev[1]{\textcolor{black}{#1}}
\newcolumntype{L}[1]{>{\raggedright\let\newline\\\arraybackslash\hspace{0pt}}m{#1}}
\newcolumntype{C}[1]{>{\centering\let\newline\\\arraybackslash\hspace{0pt}}m{#1}}
\newcolumntype{R}[1]{>{\raggedleft\let\newline\\\arraybackslash\hspace{0pt}}m{#1}}

\journal{Pattern Recognition}

\begin{document}

\begin{frontmatter}


\title{Hough-CNN: Deep Learning for \rev{Segmentation of Deep Brain Regions} in MRI and Ultrasound}


\author[1]{Fausto~Milletari\fnref{myfootnote}\corref{mycorrespondingauthor}}
\cortext[mycorrespondingauthor]{Corresponding author}
\ead[email]{fausto.milletari@tum.de}
\author[2]{Seyed-Ahmad~Ahmadi\fnref{myfootnote}}
\author[1]{Christine~Kroll}
\author[2]{Annika~Plate}
\author[2]{Verena~Rozanski}
\author[2]{Juliana~Maiostre}         
\author[2]{Johannes~Levin}
\author[3]{Olaf~Dietrich}
\author[3]{Birgit~Ertl-Wagner}
\author[2]{Kai~B\"{o}tzel}
\author[1]{Nassir~Navab}
\fntext[myfootnote]{Fausto Milletari and Seyed-Ahmad Ahmadi contributed equally to this work.}

\address[1]{Dept. of Informatics, Technische Universit\"{a}t M\"{u}nchen, Boltzmannstraße 3, Garching bei Munich, Germany}
\address[2]{Dept. of Neurology, Ludwig-Maximilians-University (LMU), Klinikum Grosshadern, Marchioninistr. 15, Munich, Germany}
\address[3]{Institute for Clinical Radiology, Ludwig-Maximilians-University (LMU), Klinikum Grosshadern, Marchioninistr. 15, Munich, Germany}

\begin{abstract}
In this work we propose a novel approach to perform segmentation by leveraging the abstraction capabilities of convolutional neural networks (CNNs). Our method is based on Hough voting, a strategy that allows for fully automatic localisation and segmentation of the anatomies of interest. This approach does not only use the CNN classification outcomes, but it also implements voting by exploiting the features produced by the deepest portion of the network. We show that this learning-based segmentation method is robust, multi-region, flexible and can be easily adapted to different modalities. In the attempt to show the capabilities and the behaviour of CNNs when they are applied to medical image analysis, we perform a systematic study of the performances of six different network architectures, conceived according to state-of-the-art criteria, in various situations. We evaluate the impact of both different amount of training data and different data dimensionality (2D, 2.5D and 3D) on the final results. We show results on both MRI and transcranial US volumes depicting respectively 26 regions of the basal ganglia and the midbrain.
\end{abstract}


\end{frontmatter}


\section{Introduction}
Recent research has shown the ability of convolutional neural networks (CNN) to deal with complex machine vision problems: unprecedented results were achieved in tasks such as classification \cite{krizhevsky12, szegedy15}, segmentation, and object detection \rev{\cite{szegedy13, sermanet13}}, often outperforming human accuracy \cite{he15}. CNNs have the ability of learning a hierarchical representation of the input data without requiring any effort to design handcrafted features \cite{lecun15}. Different layers of the network are capable of different levels of abstraction and capture different amount of structure from the patterns present in the image \cite{zeiler13}. 
Due to the complexity of the tasks and the very large number of network parameters that need to be learned during training, CNNs require a massive amount of annotated training images in order to deliver competitive results. As a consequence, significant performance increase can be achieved as soon as faster hardware and higher amount of training data become available \cite{krizhevsky12}.

In this work we investigate the applicability of convolutional neural networks to medical image analysis. Our goal is to perform segmentation of single and multiple anatomic regions in volumetric clinical images from various modalities. To this end, we perform a large study on parameter variations and network architectures, while proposing a novel segmentation framework based on Hough voting and patch-wise back-projection of a multi-atlas. We demonstrate the performance of our approach on brain MRI scans and 3D freehand ultrasound (US) volumes of the deep brain regions. 

The paradigm-shifting results delivered by CNNs in computer vision were in part accomplished with the help of extremely large training datasets and significant computational resources. Both of which may be often unrealistic in clinical environments, due to the absence of large annotated dataset and to data protection policies which often do not allow computation outsourcing.
Therefore, in this study, we perform all training and testing of CNN networks on clinically realistic dataset sizes, using a high-performance, but stand-alone PC workstation.

Segmentation of brain structures in US and MRI has widespread clinical relevance, but it is challenging in both modalities. 

\begin{figure}
	\centering
	\includegraphics[width=0.9\linewidth]{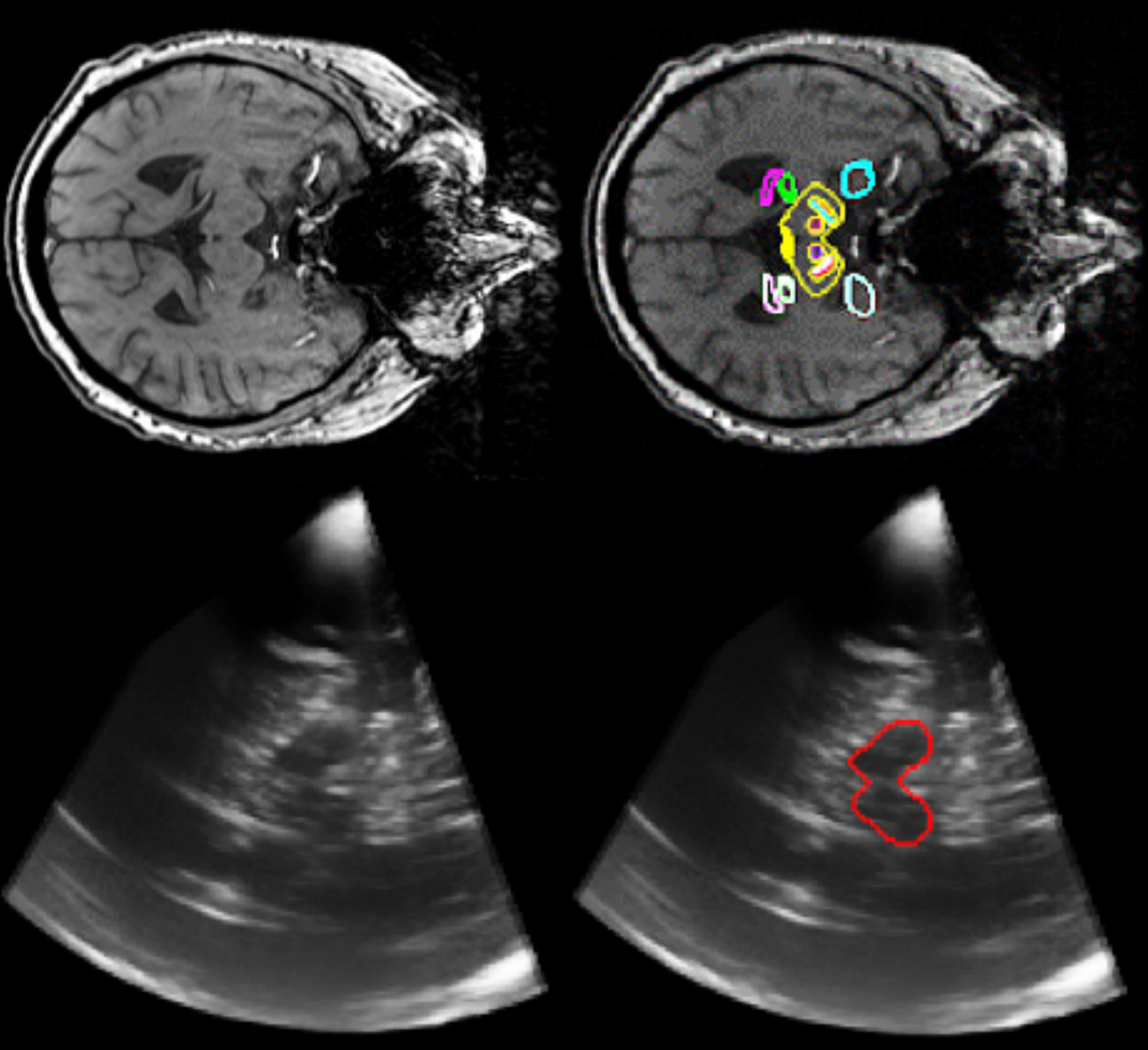}
	\caption{\rev{Example of MRI and ultrasound slices (left) and their respective segmentations (right) as estimated by Hough-CNN. Anatomies shown include midbrain in US (red) and in MRI (yellow). Further, in upper half of MRI slice: hippocampus (pink), thalamus (green), red nucleus (red), substantia nigra (green/red stripes within midbrain) and amygdala (cyan)}}
	\label{fig:MRI_fig}
\end{figure}

In MRI, the segmentation of basal ganglia is a relevant task for diagnosis, treatment and clinical research. A concrete application is pre-operative planning of Deep Brain Stimulation (DBS) neurosurgery in which basal ganglia, like the sub-thalamic nucleus (STN) and globus pallidus internal (GPi), are targeted for treatment of symptoms of Parkinson'€™s disease (PD) and dystonia, respectively \cite{dhaese12cranialvault}. Accurate localisation and outlining of these nuclei can be challenging, even when performed manually, due to their weak contrast in MRI data. Moreover, fully manual labelling of individual MRIs into multiple regions in 3D is extremely time-consuming and therefore prohibitive. For this reason, both in research \cite{dhaese12cranialvault,albis15pydbs} and in clinical practice \cite{barbe14optivise}, segmentation through atlas-based approaches is widely used.

Transcranial ultrasound (TCUS) can be used to scan deep brain regions non-invasively through the temporal bone window. Using TCUS, hyper-echogenicities of the Substantia Nigra (SN) can be analysed, gaining valuable information to perform differential \cite{walter07} and early \cite{berg11} diagnosis of Parkinson's Disease (PD). A crucial step towards computer assisted diagnosis of PD is midbrain segmentation \cite{ahmadi11, milletari15}. This task is reportedly challenging even for human observers \cite{plate12}. \rev{In order to penetrate the skull, low frequencies need to be applied resulting in an overall reduction of the resolution and in the presence of large incoherent speckle patterns. Scanning through the bone, moreover, attenuates a large part of the ultrasound energy, leading to overall reduction of the signal-to-noise ratio, as well as low contrast and largely missing contours at anatomic boundaries. Additionally, the higher speed of sound in the bone leads to phase aberration \cite{ivancevich06} and de-focussing of the ultrasound beam which causes further lowering of the image quality. A variety of image TCUS quality, anatomical visibility and 3D ultrasound fan geometry can be seen in Figure \ref{fig:SemanticVsHoughUS}. Registration methods, in particular non-linear registration, are very difficult under these conditions. Therefore, atlas-building and atlas-based segmentation methods tend to fail in ultrasound.}

In this work we evaluate the performance of our approach using an ultrasound dataset of manually annotated TCUS volumes depicting the midbrain, and an MRI dataset, depicting 26 regions including basal ganglia, annotated in a computer-assisted manner.
Our method is fully automatic, registration-free and highly robust towards the presence of artefacts. Through our patch-based voting strategy, our approach can localise and segment structures that are only partially visible or whose appearances are corrupted by artefacts. To the best of our knowledge, this is the first work employing CNNs to perform ultrasound segmentation.

Our work features several contributions:
\begin{itemize}
\item \rev{We propose Hough-CNN, a novel segmentation approach based on a voting strategy similar to \cite{milletari15}.} We show that the method is multi-modal, multi-region, robust and implicitly encoding priors on anatomical shape and appearance. Hough-CNN delivers results comparable or superior to other state-of-the-art approaches \rev{while being entirely registration-free}. In particular, it outperforms methods based on voxel-wise classification. 
\item We propose and evaluate several different CNN architectures, with varying numbers of layers and convolutional kernels per layer. In this way we acquire insights on how different network architectures cope with the amount of variability present in medical volumes and image modalities.
\item Each network is trained with different amounts of data in order to evaluate the impact of the number of annotated training examples on the final segmentation result. In particular, we show how complex networks with higher parameter number cope with relatively small training datasets.
\item We adapted the \textit{Caffe} framework \cite{jia14} to perform convolutions of volumetric data, preserving its third dimension across the whole network. We compare CNN performance using 3D convolution to the more common 2D convolution, as well as to a recent 2.5D approach \cite{roth14}.
\end{itemize}


\section{Related Works}
In this section we give an overview of existing approaches that employ CNNs to solve problems from both computer vision and medical imaging domain. 
 
In the last few years CNNs became very popular tools among the computer vision community. Classification problems such as image categorisation \cite{krizhevsky12, szegedy15}, object detection \cite{girshick2014rich} and face recognition \cite{farfade15} as well as regression problems such as human pose estimation \cite{belagiannis2015robust}, and depth prediction from RGB data \cite{eigen2014depth} have been addressed using CNNs and unprecedented results have been reported. In order to cope with the challenges present in natural images, such as scale changes, occlusions, deformations different illumination settings and viewpoint changes, these methods needed to be trained on very large annotated datasets and required several weeks to be built even when powerful GPUs were employed. In medical imaging, however, it is difficult to obtain even a fraction of this amount of resources, both in terms of computational means and amount of annotated training data. 

Many works applying deep learning to medical problems relayed only on a few dozen of training images (e.g. \cite{ciresan12,ciresan13,havaei15,ngo13,prasoon13,brebisson15}). Most networks were applied to tasks that could be solved by interpreting the images patch-wise in a sliding window fashion. In this case, several thousands of annotated training examples could be obtained from just a few images. Dataset augmentation techniques, such as random patch rotation and mirroring, were also applied if the objects of interest were invariant to these transformations \cite{roth14,ciresan12, ciresan13, havaei15}. This is the case for cell nuclei, lymph nodes and tumor regions, but not for anatomic structures with regular size and local context, such as regions of the brain or abdomen. Another way to deal with little training data is to embed CNNs as core components into previously successful methods from the community. A deep variational model is proposed in \cite{ranftl14}. Their CNN is embedded into a global inference model, i.e. the CNN outputs are treated as unary potentials on a graph and the segmentation is solved via minimum s-t cuts on the predicted graph. In \cite{turaga10} the CNN performs 3D regression to predict an affinity graph, which can be solved via graph partitioning techniques or connected components in order to segment neuron boundaries. Active shape models are realised with CNNs in \cite{liang15} via regression of multi-template contributions and object location. Variational Deep Learning was realised in \cite{ngo13} by combining shape-regularised levelset methods with Deep Belief Networks (DBN) for left ventricle segmentation in cardiac MRI.

In this work, we propose a novel Hough-CNN detection and segmentation approach. Our method utilises CNNs at its core to efficiently process medical volumes in a patch-wise fashion. It obtains voxel-wise classifications along with high level features -- used to retrieve votes -- that are descriptive of the object of interest. 
Generalised Hough voting has been proposed in the past to address problems related with object detection and tracking. Recent works such as \cite{riegler2013hough,Xie2015} performed Hough voting using a CNN. Their respective aim is to obtain head poses and cell locations in 2D by using the network to perform simultaneous classification and vote regression. In this work we propose a more flexible voting mechanism based on neighborhood relationships in feature space. On the one hand, this allows us to cast a variable amount of votes for each patch, which can be associated with information such as segmentation patches. Additionally, therapeutic indications or diagnostic information can be added or modified at any time without requiring re-training. On the other hand, instead of relying on regression, our method uses votes collected from annotated training images. Thus, it does not experience unpredictable behaviour of the votes when the network is presented with unusual data that produces unexpected feature values and mis-classifications.

Compared to computer vision which performs Deep Learning mostly on 2D images, medical images often deal with volumes acquired through scanners such as MRI or CT.  
In our literature review, most approaches have continued working in 2D by approaching 3D scans in a slice-by-slice fashion (e.g. \cite{ciresan12,ciresan13,havaei15,ngo13,prasoon13,brebisson15,kim13,lee11,song15}). The advantage is high speed, low memory consumption and the ability to utilise pre-trained nets such as AlexNet \cite{krizhevsky12}, either directly or via transfer learning. The obvious disadvantage is that anatomic context in the directions orthogonal to the image plane are entirely discarded. Some groups who employed 3D convolutions found that computational tractability was an issue, and classification was either impossible \cite{roth14} or suffered in accuracy since compromises on patch-size had to be made \cite{prasoon13}. Other groups have applied 3D convolution successfully for Alzheimer's disease detection from whole-MRI \cite{payan15} or regression of affinity graphs from 3D convolution \cite{turaga10}. A different approach that was applied to full-brain segmentation from MRI in \cite{brebisson15} combined small 3D patches with larger 2.5D ones that include more context. The 2.5D patches, in particular, consisted of a stack of three 2D patches extracted respectively from the sagittal, coronal and transversal planes. All patches were assembled into eight parallel CNN pathways in order to achieve high-quality segmentation of 134 brain regions from whole brain MRI. 

In this work, we evaluate the performance of our network when 2D, 2.5D and 3D patches are employed. In particular, we supply rather long-range 3D patches which retain a large amount of anatomical context. 

Another important issue in CNN-related research is the search for optimal CNN network architecture: we have found very little literature that addresses this issue systematically. Although several networks architectures were analysed in \cite{ciresan12, ciresan13}, we have found only one study on ``very deep CNN''  \cite{simonyan14}, in which the number of convolutional layers was varied systematically (8-16) while keeping kernel sizes fixed. The study concluded that small kernel sizes in combination with deep architectures can outperform CNNs with few layers and large kernel sizes. 

In this work we propose and benchmark six network architectures, including one very deep network having 8 convolutional layers as shown in Table \ref{table:architectures_31}.

\begin{table*}[t]
\tiny{
\begin{tabular}{|m{1.8cm}|m{5.5cm}|m{0.9cm}|m{0.7cm}|m{1cm}|}
\hline 
Name & Network Architecture & Act. function & Init. & Remarks\tabularnewline
\hline 
\hline
 3-3-3-3-3 & $I_{31} \cdot C_{3}^{64} \cdot P_{3}^{2} \cdot C_{3}^{64} \cdot C_{3}^{64} \cdot C_{3}^{64} \cdot C_{3}^{64} \cdot F_{128} \cdot F_{128} \cdot F_{\#regions}$  & \multirow{6}{*} {PReLU}   & \multirow{6}{*} {MSRA} & \multirow{6}{3cm}{$F$ use \\drop-out \\(ratio $0.5$)}\tabularnewline
\cline{1-2} 
 3-3-3-3-3-3-3-3 & $I_{31} \cdot C_{3}^{64} \cdot C_{3}^{64} \cdot C_{3}^{64} \cdot C_{3}^{64} \cdot C_{3}^{64} \cdot C_{3}^{64} \cdot C_{3}^{64} \cdot C_{3}^{64} \cdot F_{128} \cdot F_{128} \cdot F_{\#regions}$ & &  & \tabularnewline
\cline{1-2} 
 5-5-5-5-5 & $I_{31} \cdot C_{5}^{64} \cdot C_{5}^{64} \cdot C_{5}^{64} \cdot C_{5}^{64} \cdot C_{5}^{64} \cdot F_{128} \cdot F_{128} \cdot F_{\#regions}$  & & &\tabularnewline
\cline{1-2} 
7-5-3 & $I_{31} \cdot C_{7}^{64} \cdot P_{3}^{2} \cdot C_{5}^{64} \cdot C_{3}^{64} \cdot F_{128} \cdot F_{\#regions}$ & & & \tabularnewline
\cline{1-2} 
9-7-5-3-3 & $I_{31} \cdot C_{9}^{64} \cdot C_{7}^{64} \cdot C_{5}^{64} \cdot C_{3}^{64} \cdot C_{3}^{64} \cdot F_{128} \cdot F_{128} \cdot F_{\#regions}$ &   &  & \tabularnewline
\cline{1-2} 
Small Alex & $I_{31} \cdot C_{11}^{64} \cdot P_{2}^{1} \cdot C_{5}^{64}  \cdot P_{2}^{1} \cdot C_{3}^{64} \cdot C_{3}^{64} \cdot C_{3}^{64} \cdot F_{128} \cdot F_{128} \cdot F_{\#regions}$ & &  &   \tabularnewline
\hline
\end{tabular}
}
\caption{Six CNNs were designed and employed to process squared or cubic patches having size $31$ pixels. Notation for architecture and CNN layers given in section \ref{sec:ConvNets}. Activation functions follow all layers.}
\label{table:architectures_31}
\end{table*}

\section{Method} 
We propose six different convolutional neural network architectures trained with patches extracted from annotated medical volumes. We optimise our models to correctly categorise data-points into different classes. The volumes were acquired in two different modalities, US and MRI, and depict deep structures of the human brain. Accurate segmentation of the desired regions has been achieved through a Hough voting strategy, inspired by \cite{milletari15}, which was employed to simultaneously localise and segment the structures of interest.

\subsection{Convolutional neural networks}
\label{sec:ConvNets}
A CNN consists of a succession of layers which perform operations on the input data. \textit{Convolutional layers} (symbol $C_s^k$) convolve the images $I_{size}$ presented to their inputs with a predefined number ($k$) of kernels, having a certain size $s$, and are usually followed by \textit{activation units} which rescale the results of the convolution in a non linear manner. \textit{Pooling layers} (symbol $P_{size}^{stride}$) reduce the dimensionality of the responses produced by the convolutional layers through downsampling, using different strategies such as average-pooling or max-pooling. Finally, \textit{fully connected layers} (symbol $F_{\#neurons}$) extract compact, high level features from the data. The kernels belonging to convolutional layers as well as the weights of the neural connections of the fully connected layers are optimised during training through back-propagation. The network architecture is specified by the user, by defining the number of layers, their kind, and the type of activation unit. Other relevant parameters are: the number and size of the kernels employed during convolution, the amount of neurons in the fully connected part and the downsampling ratio applied by the pooling layers. We propose six network architectures that are described in Table \ref{table:architectures_31}.

CNNs perform machine learning tasks without requiring any handcrafted feature to be engineered and supplied by the user. That is, discovering optimal features describing the data at hand is part of the learning process. 
During training the network parameters are first initialised and then the data is processed through the layers in a feed-forward manner. The output of the network is compared with the ground-truth through a loss function and the error is back-propagated \cite{lecun15} in order to update the filters and weights of all the layers, up to the inputs. This process is repeated until it converges. Once the network is trained, predictions can be made by using it in a feed-forward manner and reading out the outputs of the last layer.

In our approach we made use of parametric rectified linear units \cite{he15} (PReLU) as our activation functions.
\begin{equation}
PReLU(x)=\begin{cases}
x \mbox{ if } x \geq 0 \\
\alpha x \mbox{ if } x < 0\end{cases}
\end{equation}
The parameter $\alpha$ in the PReLU activation function is learnt during training, along with other network weights. In this context we initialise the network parameters using MSRA \cite{he15} as it is an appropriate choice when employing PReLU activation units.

Many authors \cite{krizhevsky12, hinton2012improving} reported that the tendency of the network to overfit can be decreased by using a technique called ``drop-out'' during training which inhibits the outputs of a random fraction of the neurons of the fully connected layers in each iteration. In this way it is possible to limit their excessive specialisation to specific tasks, which is believed to be at the origin of overfitting in CNNs.

Finally, we employ max-pooling layers to reduce the dimensionality of the data as it traverses the network. The input of the pooling layer is exhaustively subdivided into sub-patches having fixed size and overlapping by an amount controlled by the ``stride'' parameter. Only the maximal value in each sub-patch is forwarded to the next layer. This procedure is known to incorporate a spatial invariance to the network which contradicts the desired localisation accuracy required for segmentation. For this reason we limit the usage of pooling layers to the minimum amount required to meet the existing hardware constraints.

\subsection{\rev{Voxel}-wise classification}
\begin{figure*}
	\centering
	\includegraphics[width=1.0\linewidth]{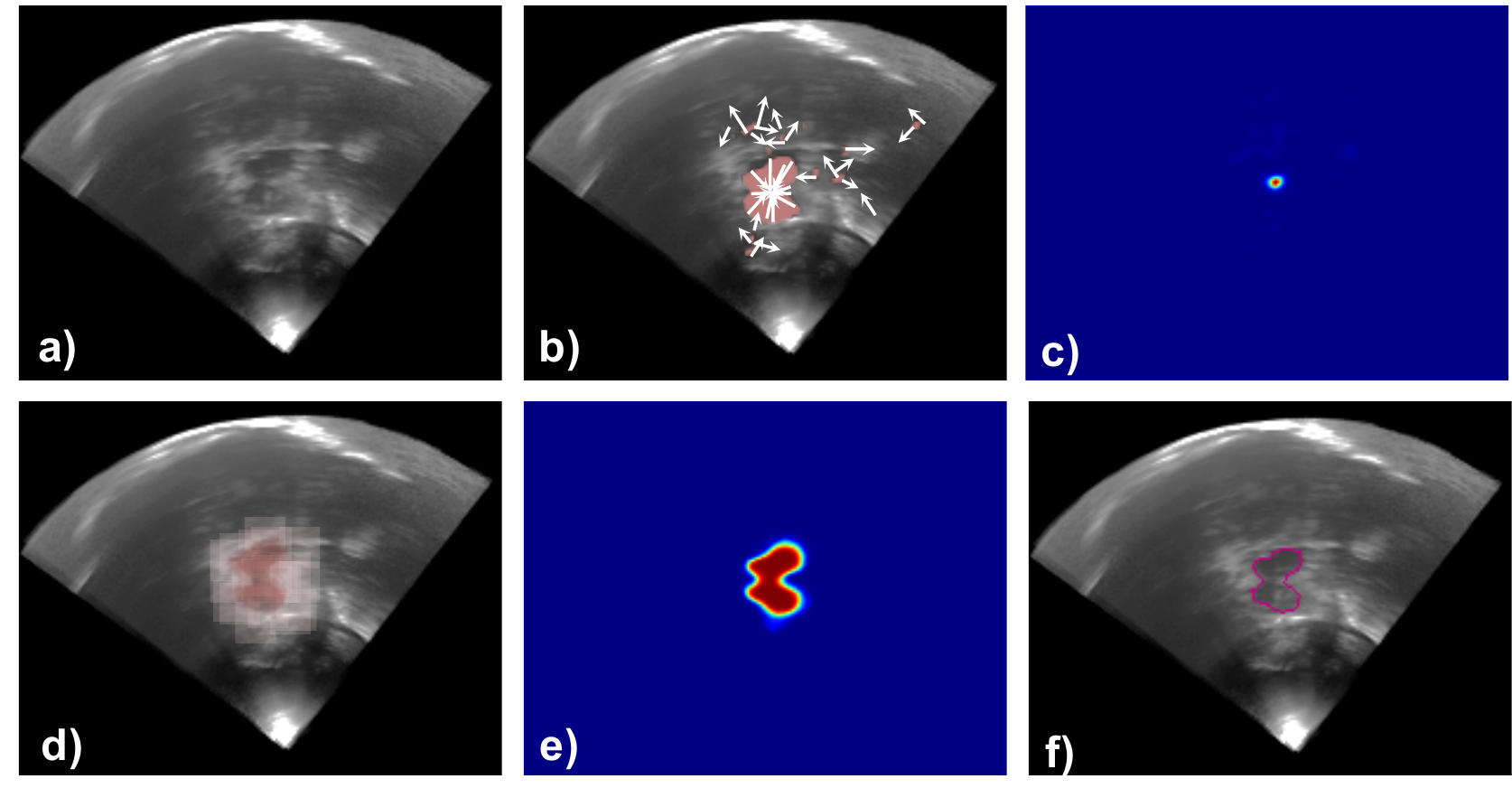}
	\caption{\rev{Schematic representation in 2D of the Hough-CNN segmentation approach. a) The volume is interpreted patch-wise and classified using the CNN. b) Every pixel of the foreground (red) casts one or multiple votes in order to localise the anatomy centroid. c) The votes accumulate in a vote-map, represented here in jet colormap, and the object centroid is found at the location of maximum vote accumulation. d) All the votes that accumulated close to the detected anatomy centroid contribute to the final contour by projecting a binary segmentation patch  (here shown in red and white to indicate foreground and background respectively) at the location they were cast from. e) A contour confidence map is constructed by accumulating all the contributions associated to the votes. f) The resulting contour, depicted in purple, is retrieved by thresholding the confidence map.}}
	\label{fig:Hough-CNN_explained}
\end{figure*}

A set $\mathbf{T}= \{ \mathbf{p}_1, ... , \mathbf{p}_N \}$ of square (or cubic) patches having size $p$ pixels is extracted from $J$ annotated volumes $V_j$ with $\mbox{ }j\in\left\{ 1...J\right\}$ along with the corresponding ground truth labels  $\mathbf{Y}= \{ y_1, ... , y_N \} \in \mathbb{R}$. Based on this training set CNNs are optimised to categorise the patches correctly. The resulting trained networks are capable of performing \rev{voxel}-wise classification, \rev{also called semantic segmentation}, of volumes by interpreting them in a patch-wise fashion. However, due to the lack of regularisation and enforcement of statistical priors this approach delivers sub-optimal results \rev{(Figure \ref{fig:MRI_pixelwise})}. For this reason we introduce a novel segmentation method that is based on simultaneous localisation of the anatomy of interest and robust contour extraction \rev{(Figure \ref{fig:Hough-CNN_explained})}.

\subsection{Hough voting with CNN}
We introduce a robust segmentation approach that is scalable to multiple regions and implicitly encodes shape priors. This method employs a Hough-voting strategy to perform anatomy localisation and a database containing segmentation patches to retrieve the contour of the anatomy. 
Instead of relying only on categorical predictions produced by the CNNs we also make use of features extracted from their intermediate layers, in particular from the second-last fully connected one. Several authors \cite{krizhevsky12,girshick2014rich,farfade15} have reported that these features (sometimes also called descriptors) can be used for tasks such as image retrieval by mapping images to the feature space and identifying their neighbours. These findings are employed at the core of our voting strategy. 

To keep our notation as simple and understandable as possible we describe our approach for single region segmentation in the following.

\rev{
During \textit{training}, we make use of the dataset of training volumes $\mathbf{V}_{j}$ with $\mbox{ }j\in\left\{ 1...J\right\}$, and respective binary segmentation volumes $\mathbf{S}_{j}$ with $\mbox{ }j\in\left\{ 1...J\right\}$. We collect patches from both foreground and background and train a CNN. As a result, we obtain the parameters $\hat{\theta}$ that define the network. 
The CNN not only differentiates patches belonging to foreground and background through classification, but also associates each input to a feature vector obtained from its second-last fully connected layer. The macroscopic effect of the network can be summarised using two functions
}
\[
\rev{f_1(\mathbf{p}_{i},\hat{\theta})=l_{i}\in\{0,1\}\mbox{ and }\mathbf{f}_2(\mathbf{p}_i,\hat{\theta})=\mathbf{f}_{i}\in\mathbb{R}^{d}}
\]
\rev{respectively  mapping each input  patch $\mathbf{p}_i$ to its label $l_i$ and to the feature $\mathbf{f}_i$, which has as many dimensions $d$ as there are neurons in the fully connected layer it is collected from.}

\rev{We exhaustively collect a dataset $\mathbf{T}=\left\{ \mathbf{p}_1...\mathbf{p}_N\right\}$ of either $2$D, $2.5$D or $3$D patches from the locations $\mathbf{X}=\left\{\mathbf{x}_{1}...\mathbf{x}_{N}\right\}$ of the foreground region of each of the training volumes $\mathbf{V}_j$, and we use the CNN to obtain the features $\mathbf{f}_i$ introduced before.
Our goal is to create a database storing triples consisting of a feature vector $\mathbf{f}_i$, a vote $\mathbf{v}_i$ and a segmentation patch $\mathbf{s}_i$.}

\rev{The vote $\mathbf{v}_{i}$ is a displacement vector joining the voxel $\mathbf{x}_{i}$, where the $i$-th patch was collected from, and the position anatomy centroid $\mathbf{c}_j$ in the training volume $\mathbf{V}_j$:}
\[
\rev{\mathbf{v}_{i}=\mathbf{x}_{i}-\mathbf{c}_{j};\;\mathbf{c}_{j}=\frac{1}{|F_g|}\sum_{\mathbf{x}_i\in F_g}\mathbf{x}_i}
\]

\rev{where $F_g$ is the set of all the voxels belonging to foreground. The binary segmentation patches assume values $1$ or $0$ respectively for foreground and background area since they are collected from the positions $\mathbf{x}_{i}$ of the binary annotation volumes $\mathbf{S}_j$.}
 



\rev{During \textit{testing}, in order to segment a previously unseen volume $I$, we make use of both the trained CNN and the database established before.}
We first obtain \rev{the classification label} for each voxel \rev{$\mathbf{x}_i$} by processing the relative patch \rev{$\mathbf{p}_i$} through the CNN, \rev{which delivers also the} features $\mathbf{f}_i$ for all the patches being classified as foreground. Each of such features is compared to those contained in the \rev{database} in order to retrieve the $K$ closest entries using \rev{Euclidean distance as criterion}. This $K$-nearest neighbour search ($K$-nn) \cite{flann_pami_2014} is performed computing Euclidean distances $d^i_{1...K}$ between features, as previously done in \cite{krizhevsky12} for image retrieval. 

Once the neighbours are identified, \rev{their} votes \rev{$\mathbf{v}^i_{1...K}$ and associated segmentation patches $\mathbf{s}^i_{1...K}$ from the database, are employed to respectively perform localisation and segmentation}. The votes are weighted by the \rev{reciprocal} of the Euclidean distance \rev{computed} during $K$-nn search $w_{1...K}=\frac{1}{d^i_{1...K}}$ and contribute to a vote-map at positions 
\[
\hat{\mathbf{v}}^i_k=\mathbf{x}_i+\mathbf{v}^i_k;\mbox{ }\forall k\in\{1...K\}
\]
\rev{We repeat the steps described above for each of the patches that were classified as foreground (Figure \ref{fig:Hough-CNN_explained}b).} Since the region of interest occurs only once in each volume, we smooth the final vote map and retrieve the region centroid by finding the location \rev{$\mathbf{c}$} where the maximal value of the vote map is reached (Figure \ref{fig:Hough-CNN_explained}c). \rev{Smoothing reduces the possibility of small localisation mistakes due to ``noise'' in the vote map around the position where its maximum occurs.} 

\rev{The region of interest can now be segmented by re-projecting the votes $\mathbf{v}^i_k$ to the locations $\mathbf{x}_i$ where they have been originated from}. \rev{However, not all the votes should be re-projected, since a relevant portion of them is erroneous, i.e. did not contribute to the vote-map anywhere close to the estimated anatomy location. Thus, only those that contributed to the vote-map within a certain range $r$ from the predicted centroid are taken into consideration and are actually allowed to contribute to the final segmentation contour with their own segmentation patch $\mathbf{s}^i_k$. The segmentation patches $\mathbf{s}^i_k$ are centred at the location $\mathbf{x}_i$, weighted by $w^i_k$ and accumulated in the segmentation map $\mathbf{S}$ (Figure \ref{fig:Hough-CNN_explained}d).} \rev{Assuming that the segmentation patches $\mathbf{s}^i_k$ have been extended to an infinite spatial extent by zero-padding, we can write:
\[
\hat{\mathbf{S}}(\mathbf{x})=\sum_{\mathbf{x}_{i}}\sum_{k=1}^{K}Ind(\hat{\mathbf{v}}_{k}^{i},\hat{\mathbf{c}})\mbox{ }w_{k}^{i}\mbox{ }s^i_{k}(\mathbf{x}-\mathbf{x}_{i})
\]
\[Ind(\mathbf{a},\mathbf{b})=\begin{dcases}
1 & \left\Vert \mathbf{a}-\mathbf{b}\right\Vert <r \\
0 & \left\Vert \mathbf{a}-\mathbf{b}\right\Vert \geq r
\end{dcases}
\]}
\rev{In this sense, the segmentation patches $\mathbf{s}^i_k$ can be seen as basis functions $s^i_k(\mathbf{x})$, which take binary values, that need to be scaled and re-centered at appropriate locations in order to produce the desired effect in the segmentation map.}
Once the segmentation map $\mathbf{S}$ is normalised to take only values comprised between $0$ and $1$, it is thresholded and the final contour is obtained.

\rev{The approach is summarised schematically in Figure \ref{fig:Hough-CNN_explained}}.
Extending this method to multiple regions requires little effort. In our implementation, we treated each region independently by creating region-specific databases as well as dedicated vote-maps and segmentations. 
\rev{The memory requirements of this approach can be decreased by retrieving the segmentation patches directly from the volumes $\mathbf{S}_{1...J}$ instead of storing them in the database. In this case, the database contains coordinates that are used to fetch contour portions from the $\mathbf{S}_{1...J}$.}

\subsection{Efficient patch-wise evaluation through CNN}
When dealing with images or volumes, patches are extracted in a sliding-window fashion and processed through a CNN. This approach is inefficient due to the high amount of redundant computations that need to be performed for neighbouring patches. In case no padding is used within the convolutional layers, the whole volume can be convolved with the respective kernels in one pass, instead of treating each patch separately, while achieving the same result. The same holds true for pooling layers whose pooling windows can be arranged to process the whole volume at once. However, as soon as fully connected layers are employed, the volume can no longer be processed in one pass due to the fact that the connections of this layer are limited to the size of the input patch.

\rev{To solve this issue we modify the network structure as proposed by Sermanet et al. in \cite{sermanet2013overfeat} in order to be able to process the whole volume at once, yet retrieving the same results that we would obtain if the data would be processed patch-wise.}

\begin{figure}
	\centering
	\includegraphics[scale=0.7]{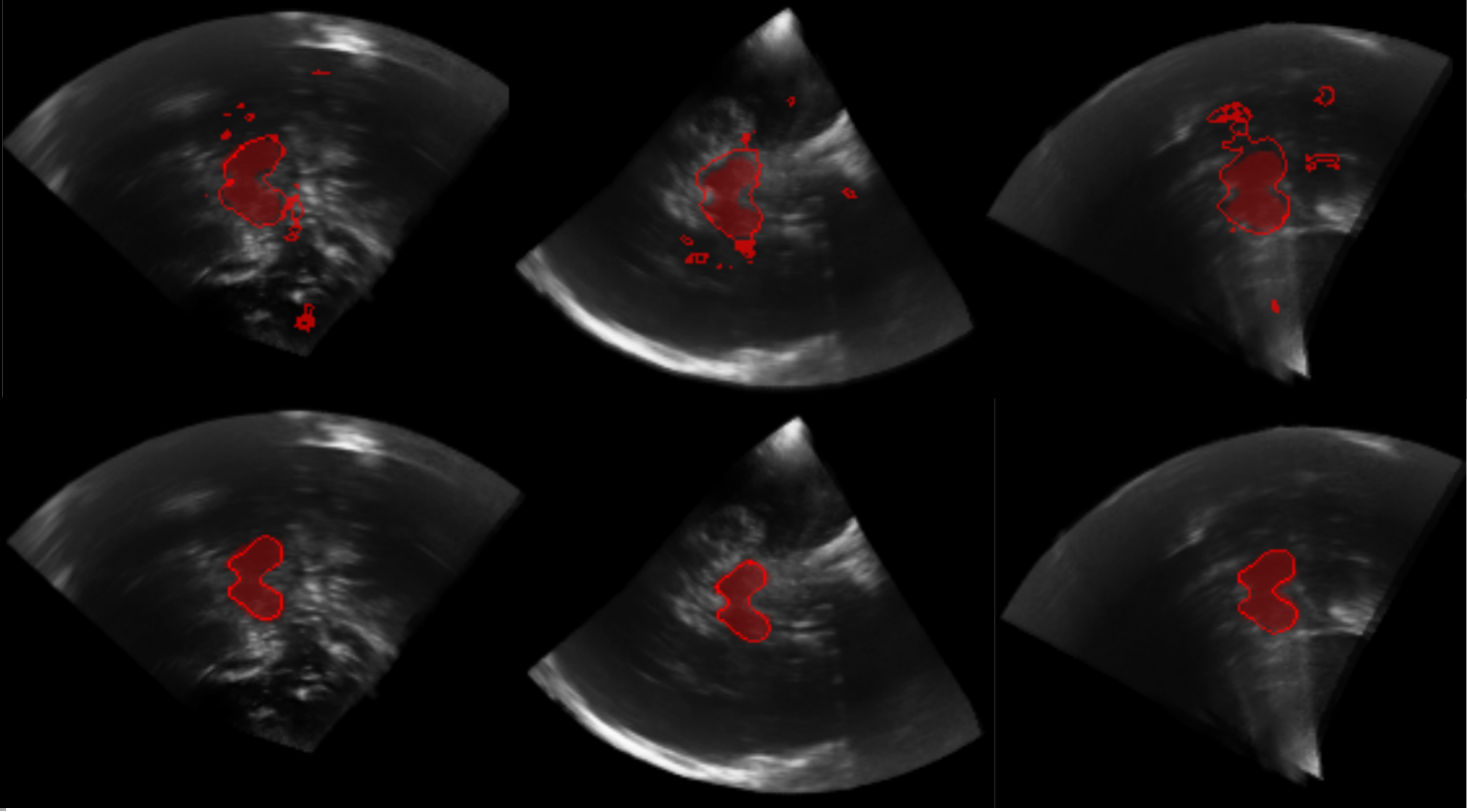}
	\caption{\rev{Visual comparison of semantic segmentation results (top) and Hough-CNN results (bottom) on the same ultrasound data using the best-performing CNN. Red areas represent ground truth annotation. Red contours represent segmentation outputs. Best viewed in digital format.}}
	\label{fig:SemanticVsHoughUS}
\end{figure}

\section{Experiments and Results}
\begin{figure}
	\centering
	\includegraphics[scale=0.7]{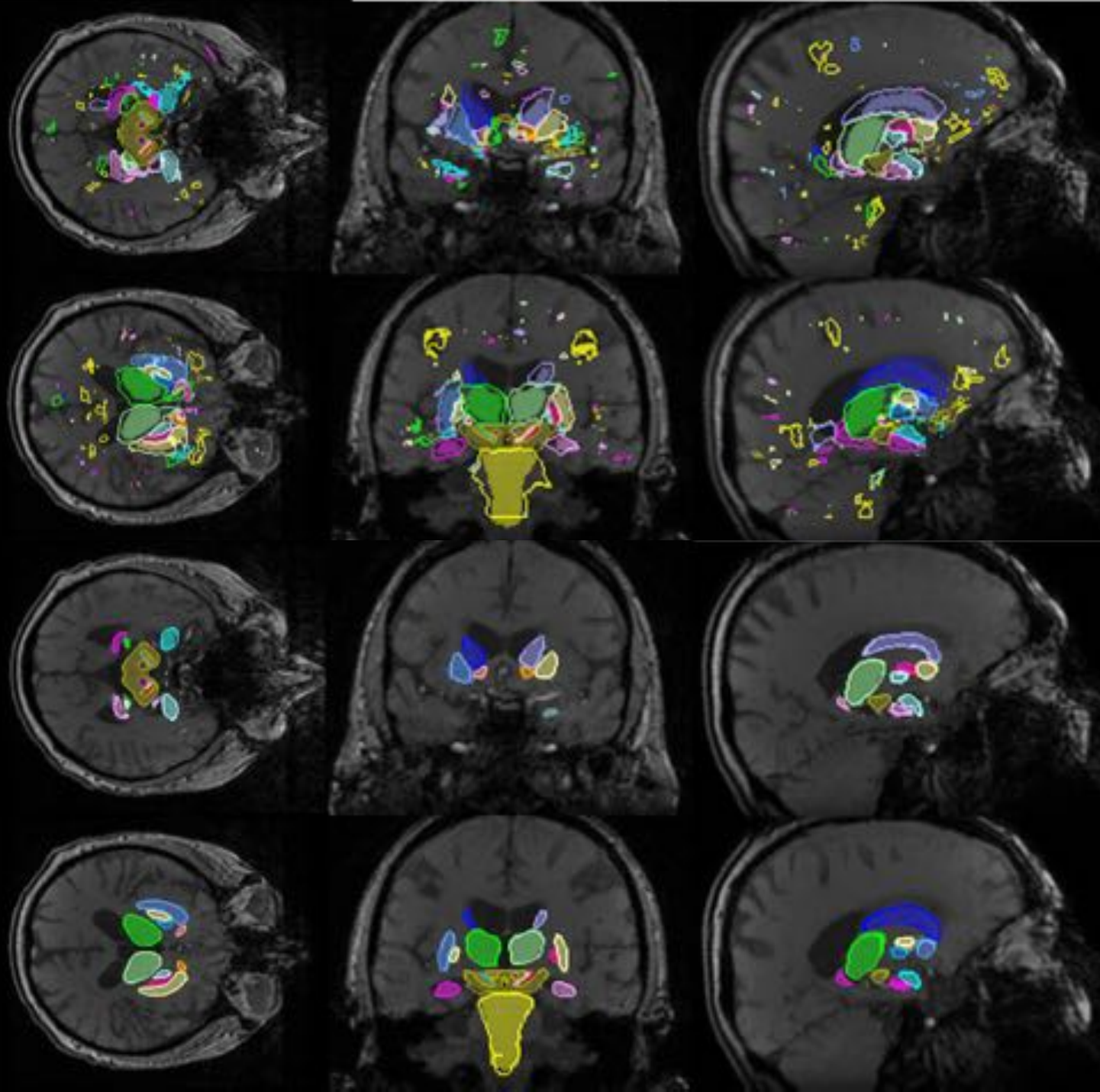}
	\caption{\rev{Visual comparison of semantic segmentation results (top two rows) and Hough-CNN results (bottom two rows) on same MRI volumes using the same trained CNN. Coloured areas represent ground truth annotation. Coloured contours represent segmentation outputs. Best viewed in digital format.}}
	\label{fig:SemanticVsHoughMRI}
\end{figure}

In this section we show that CNNs not only can be used to robustly segment medical volumes \rev{(Figure \ref{fig:SemanticVsHoughUS}, Figure \ref{fig:SemanticVsHoughMRI})}, but they also posses the ability of learning extremely effective features \rev{(outputs of upper layers)} from the data. Even in ultrasound, where the structures of interest are often not clearly visible or the images are affected by artefacts, CNNs are able to focus on salient information and therefore recognise patterns.
We demonstrate the superior performances of our Hough-voting-based segmentation algorithm by evaluating our method on two datasets of US and MRI volumes depicting the human brain. The two modalities provide complementary information, but are inherently different both from the point of view of the challenges they offer and the range of anatomy they can image. 

\subsection{Datasets and ground-truth definition}
\label{sec:datasets}
Our MRI dataset is composed of MRI volumes of $55$ subjects, which were acquired using 3D gradient-echo imaging (magnitude and phase) with an isotropic spatial resolution of 1x1x1 mm. The sequence \cite{dietrich15} is designed for quantitative susceptibility mapping (QSM) and sensitivity towards iron deposits. These are biomarkers for movement disorders like Parkinson's Disease and create visible contrast in relevant basal ganglia like SN and STN. 
For our study, basal ganglia and other deep-brain structures were annotated in an atlas volume \rev{in two ways. One set of bi-lateral atlas labels (brainstem, n. accumbens, amygdala, caudate, thalamus, hippocampus, pallidum, putamen) were annotated semi-automatically via a shape- and appearance-model segmentation (FSL FIRST \cite{patenaude11}) plus manual correction of generated labels (one neuroimage technician, verified by one expert neurologist). Another set of bi-lateral labels (separation of of pallidus into GPi and GPe, midbrain, red nucleus, substantia nigra pars compacta and substantia nigra pars reticulata) was annotated in a fully manual manner (neuroimage technician, verified by expert neurologist) based on visible contrast. The atlas labels were transferred using a state-of-the-art atlas approach \cite{avants10}.} As a summary, the list of structures of interest is also visible in Figure \ref{fig:MRI_biggest}.  


The US dataset was acquired transcranially on $34$ subjects, with several freehand 3D sweeps recorded through the left and right temporal bone window each. Altogether, $162$ volumes were acquired with slight variations in bone window positioning, and reconstructed at 1mm isotropic resolution. 
For all $162$ TCUS volumes, midbrain outlines were annotated in 3D by a single human expert. \rev{Inter-rater agreement of the midbrain annotations, in terms of Dice coefficient, has been reported in \cite{plate12} to be $0.85$}. CNN training was performed on data from $8$ subjects ($40$ sweeps), and testing on data from $24$ previously unseen subjects ($114$ sweeps), while validation data was performed on 8 sweeps from $2$ subjects. \rev{Performing segmentation on more than 100 test volumes is a good indicator of actual clinical applicability of (Hough-)CNN-based segmentation. The experiments show that the method generalises very well on previously unseen data, which is a highly desirable property in clinical settings.}

In order to test our approach and to benchmark the capabilities of the proposed CNNs when they are trained with a variable amount of data, we establish, for each dimensionality (2D, 2.5D and 3D) two differently sized training sets in US and three in MRI respectively. 
For each of the $40$ training volumes in US we collect either $2$K or $10$K patches per volume such that half of the training set depicts the background and the other half the foreground. The resulting training sets have respective sizes of $80$K and $400$K patches. A validation set containing $5$K patches has been established for US using images of subjects that have not been used for training or testing and employed to assess the generalisation capabilities of the models.
From the $45$ MRI training volumes, we extract either circa $100$, $1$K  or $10$K patches per volume \textit{per region} (including background). The resulting training sets have respective sizes of $135$K, $1.35$M and $13.5$M patches.

\subsection{CNN parameters}
\label{sec:CNNparameters}


We analyse six different network architectures, presented in Table \ref{table:architectures_31}, by training each of them for 15 epochs using Stochastic Gradient Descent (SGD) with mini-batches of 64 or 124 samples, learning rate varying between $10^{-2}$ and $5\cdot10^{-3}$ depending on the individual network architecture, momentum $0.9$ and weight decay $5\cdot10^{-4}$. All our models converged after a few epochs, and often before the seventh epoch.

Each network is analysed three times, with patches capturing the same amount of context from the neighbourhood, but having different dimensionality. That is, our networks process 2D data, 2.5D data and 3D data in order to investigate how the networks respond to the higher amount of information carried by patches in 2.5D and 3D patches compared to 2D. During training, we randomly sample patches from annotated volumes and we feed them to the networks along with their ground truth labels. The patches of the $2D$ dataset are all square and have a size of $31\times31$ pixels; the 2.5D dataset is composed of patches having the same size and three channels consisting of 2D patches from the sagittal, coronal and transversal plane centred at the same location; the 3D dataset contains cubic patches having size $31\times31\times31$ voxels.

Some of the parameters supplied to our Hough-CNN algorithm are empirically chosen. Parameters names and respective values are reported in Table \ref{table:parameters}. These parameters remained constant throughout all experiments, both in ultrasound and MRI. All the trainings were performed on Intel i7 quad-core workstations with 32 gigabytes of ram and graphic cards from Nvidia, specifically "Tesla k40" or "Titan X" (12GB VRAM). All tests were made on a similar workstation equipped with a Nvidia GTX 980 (4 GB VRAM).

\begin{table}
\centering
\begin{tabular}{|c|c|}
\hline 
Parameter Name & Value\tabularnewline
\hline 
\hline 
Tolerance radius $r$ for reprojection & $r=3$ voxels\tabularnewline
\hline 
Amount of smoothing for vote-maps & $\sigma=1$\tabularnewline
\hline 
Maximum number of neighbours K-NN & $K=20$\tabularnewline
\hline 
Maximal distance of K-NN neighbours (US) & $2.5$\tabularnewline
\hline
Maximal distance of K-NN neighbours (MRI) & $6.0$\tabularnewline
\hline 
Size of segmentation patch & $9\times9\times9$\tabularnewline
\hline 
\end{tabular}
\caption{Parameters of the model utilised during the experiments.}
\label{table:parameters}
\end{table}

\begin{sidewaystable}

\begin{tabular}{|c|c|c|c|c|c|c|c|c|c|}
\hline 
Dimensionality $\rightarrow$ & \multicolumn{3}{c|}{2D} & \multicolumn{3}{c|}{2.5D} & \multicolumn{3}{c|}{3D}\tabularnewline
\hline 
Averages $\rightarrow$ & Dice $[0,1]$ & Distance (mm) & Failures & Dice $[0,1]$ & Distance (mm) & Failures & Dice $[0,1]$ & Distance (mm) & Failures\tabularnewline
\hline 
\hline 
\multicolumn{10}{|c|}{Training set size $80$K patches}\tabularnewline
\hline 
3-3-3-3-3 & \textbf{0.83} & 0.92 & 3\% & \underline{0.82} & 0.91 & 5\% & 0.79 & 0.95 & 6\%\tabularnewline
\hline 
3-3-3-3-3-3-3-3 & 0.80 & 0.93 & 5\% & 0.80 & 0.94 & 4\% & \underline{\textbf{0.82}} & 0.99 & 5\%\tabularnewline
\hline 
5-5-5-5-5 & 0.77 & 1.07 & 9\% & 0.74 & 1.11 & 14\% & \textbf{0.80} & 1.02 & 6\%\tabularnewline
\hline 
7-5-3 & 0.80 & 0.96 & 5\% & \textbf{0.81} & 1.00 & 5\% & 0.80 & 1.02 & 7\%\tabularnewline
\hline 
9-7-5-3-3 & 0.79 & 0.96 & 7\% & 0.81 & 0.93 & 5\% & \underline{\textbf{0.82}} & 0.99 & 7\%\tabularnewline
\hline 
SmallAlex & \underline{\textbf{0.85}} & 0.81 & 1\% & 0.81 & 0.98 & 5\% & 0.80 & 0.98  & 3\%\tabularnewline
\hline 
\hline 
\multicolumn{10}{|c|}{Training set size: $400$K patches}\tabularnewline
\hline 
3-3-3-3-3 & \textbf{0.84} & 0.90 & 1\% & \underline{0.83} & 0.95 & 3\% & \multirow{6}{*}{--} & \multirow{6}{*}{--} & \multirow{6}{*}{--}\tabularnewline
\cline{1-7} 
3-3-3-3-3-3-3-3 & \underline{\textbf{0.85}} & 0.90 & 0\% & \underline{0.83} & 0.99 & 3\% &  &  & \tabularnewline
\cline{1-7} 
5-5-5-5-5 & \textbf{0.83} & 0.94 & 2\% & 0.81 & 1.03 & 5\% &  &  & \tabularnewline
\cline{1-7} 
7-5-3 & \textbf{0.83} & 0.94 & 2\% & 0.81 & 0.99 & 5\% &  &  & \tabularnewline
\cline{1-7} 
9-7-5-3-3 & \textbf{0.82} & 1.01 & 2\% & \textbf{0.82} & 0.96 & 5\% &  &  & \tabularnewline
\cline{1-7} 
SmallAlex & \textbf{0.83} & 0.91 & 3\% & 0.81 & 0.94 & 4\% &  &  & \tabularnewline
\hline 
\end{tabular}
\label{table:USresults}
\caption{Midbrain segmentation results in 114 previously unseen TCUS volumes, using Hough-CNN with variations of architectures (single rows), patch dimensionalities (column blocks) and training set sizes (row blocks). \rev{The best result for each architecture (across the data dimensionalities) are highlighted by using bold typeface. The best results for each dimensionality (across the architectures) are underlined.}}
\end{sidewaystable}

\subsection{Experiments and results in ultrasound}
\label{sec:ResultsUltrasound}
We train our CNNs with different amount of data having different dimensionality, as explained in Section \ref{sec:datasets}. Each of the six proposed architectures is trained six times (five for 3D) in order to cover all the possible combinations of dimensionalities ($2$D, $2.5$D, $3$D patches) and amount of data (training set sizes $80$K, $400$K). We test each CNN on $114$ ultrasound volumes acquired from subjects whose scans have never been used during training or validation. 

\begin{figure*}
	\centering
 	\includegraphics[scale=0.3]{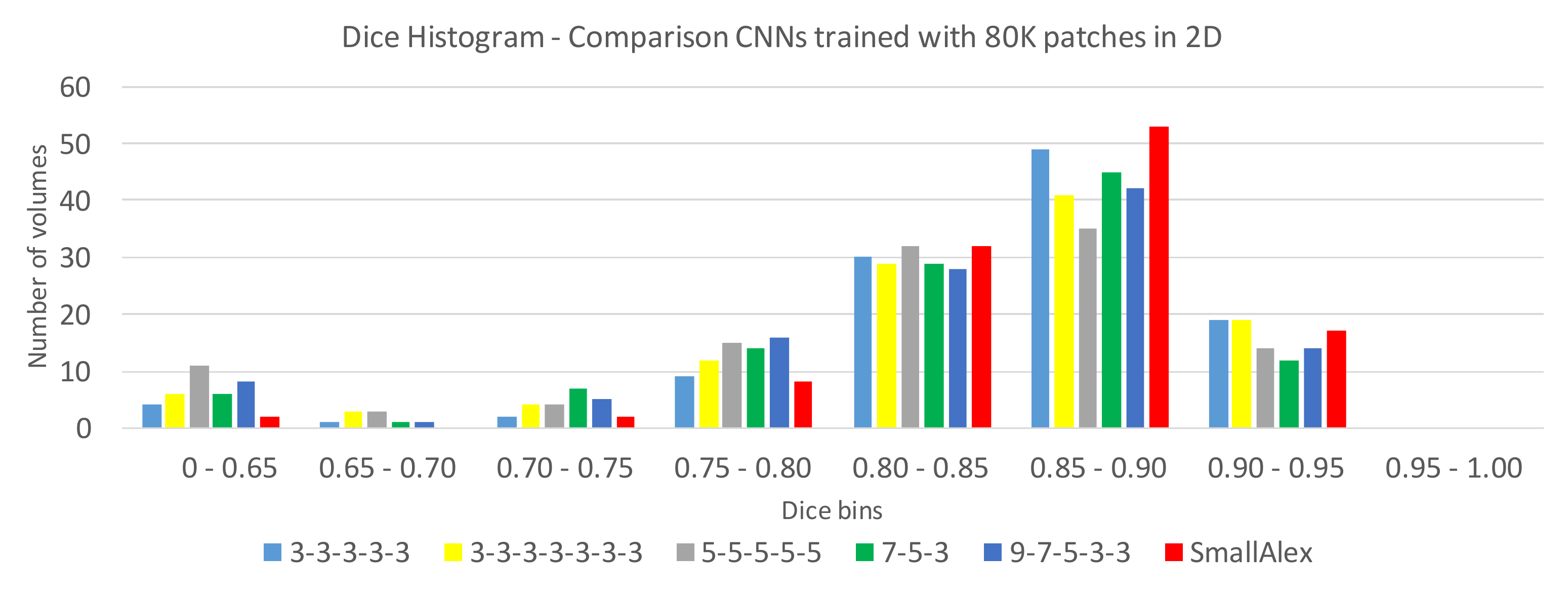}
	\includegraphics[scale=0.3]{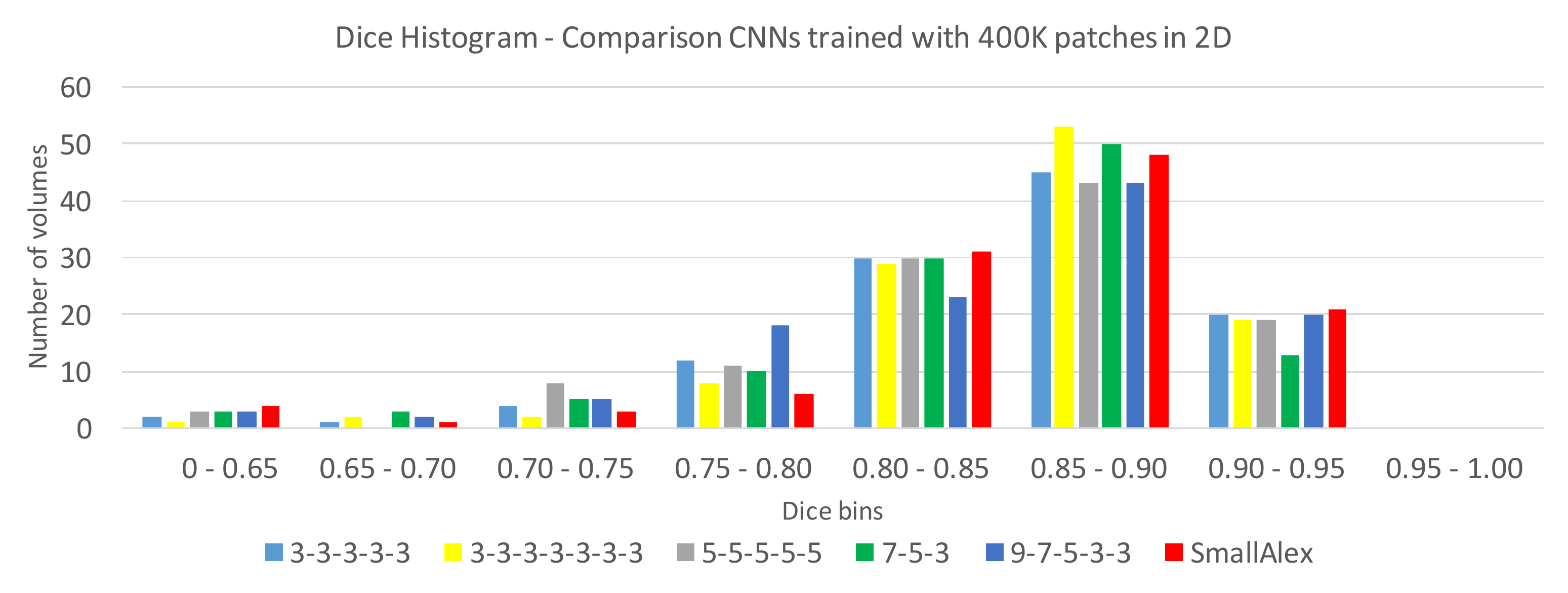}
 	\includegraphics[scale=0.3]{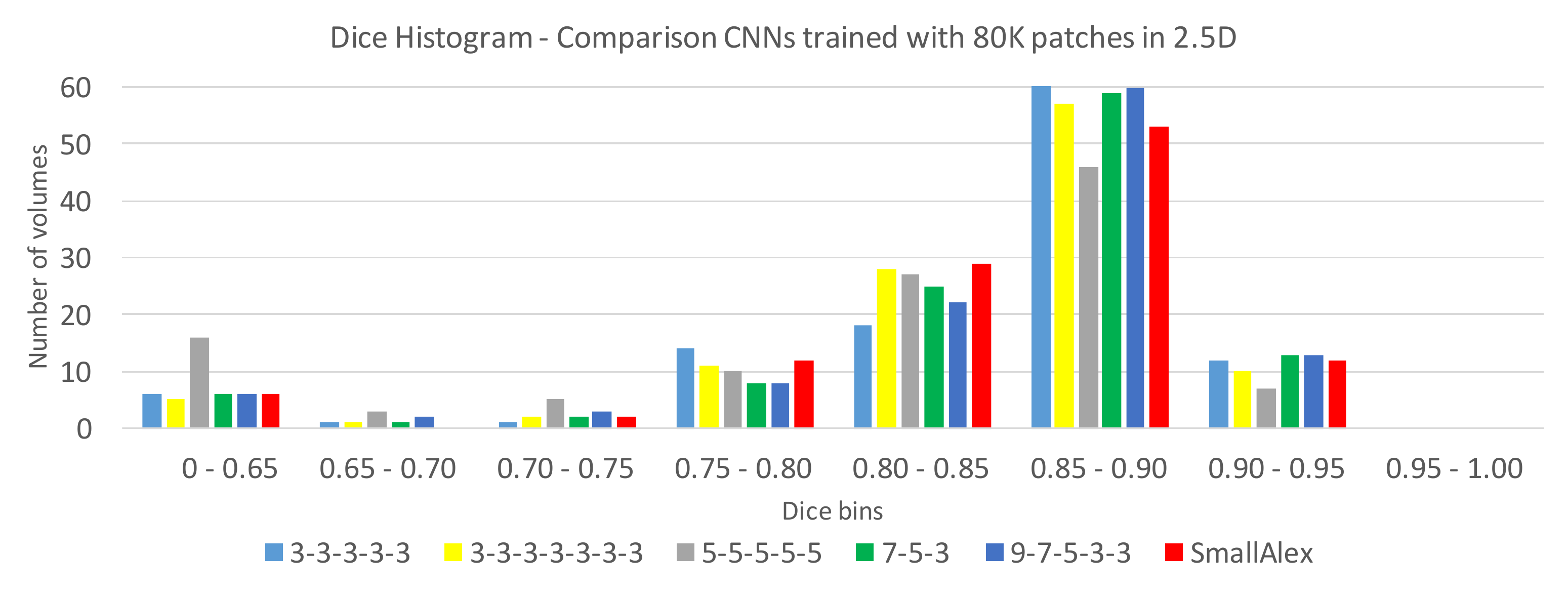}
	\includegraphics[scale=0.3]{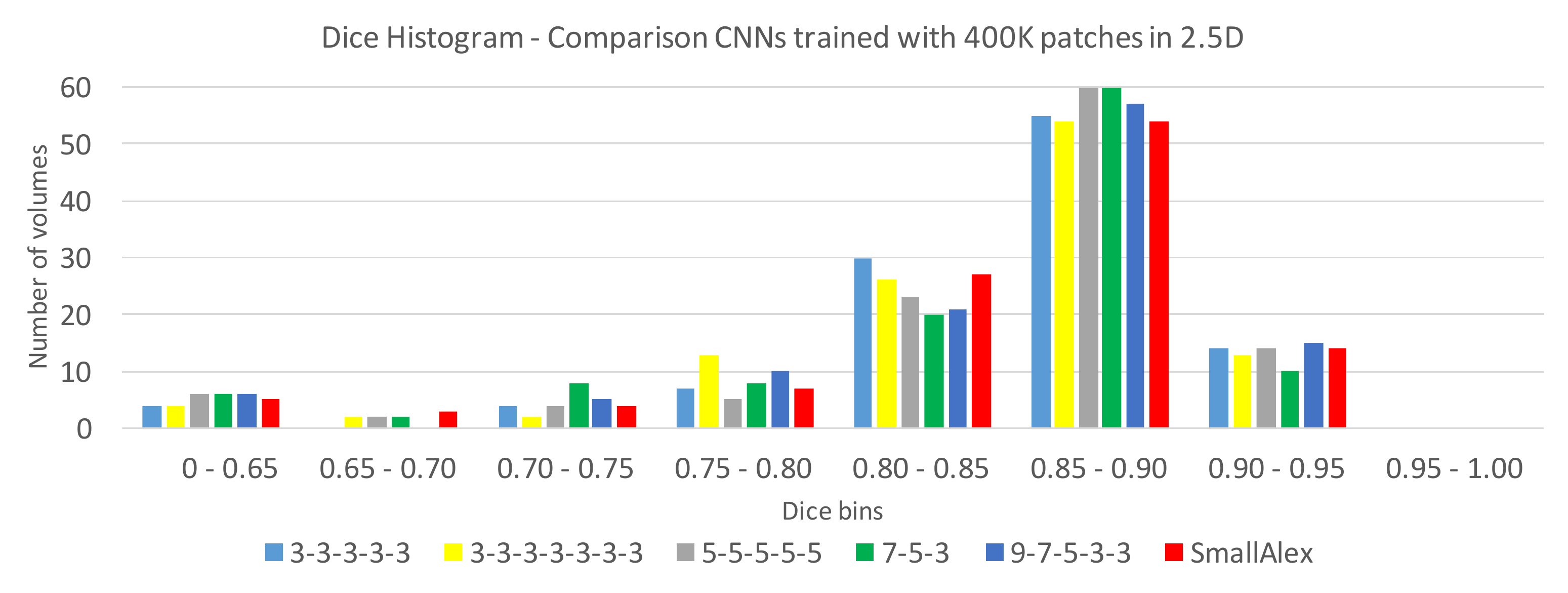} 
 	\includegraphics[scale=0.3]{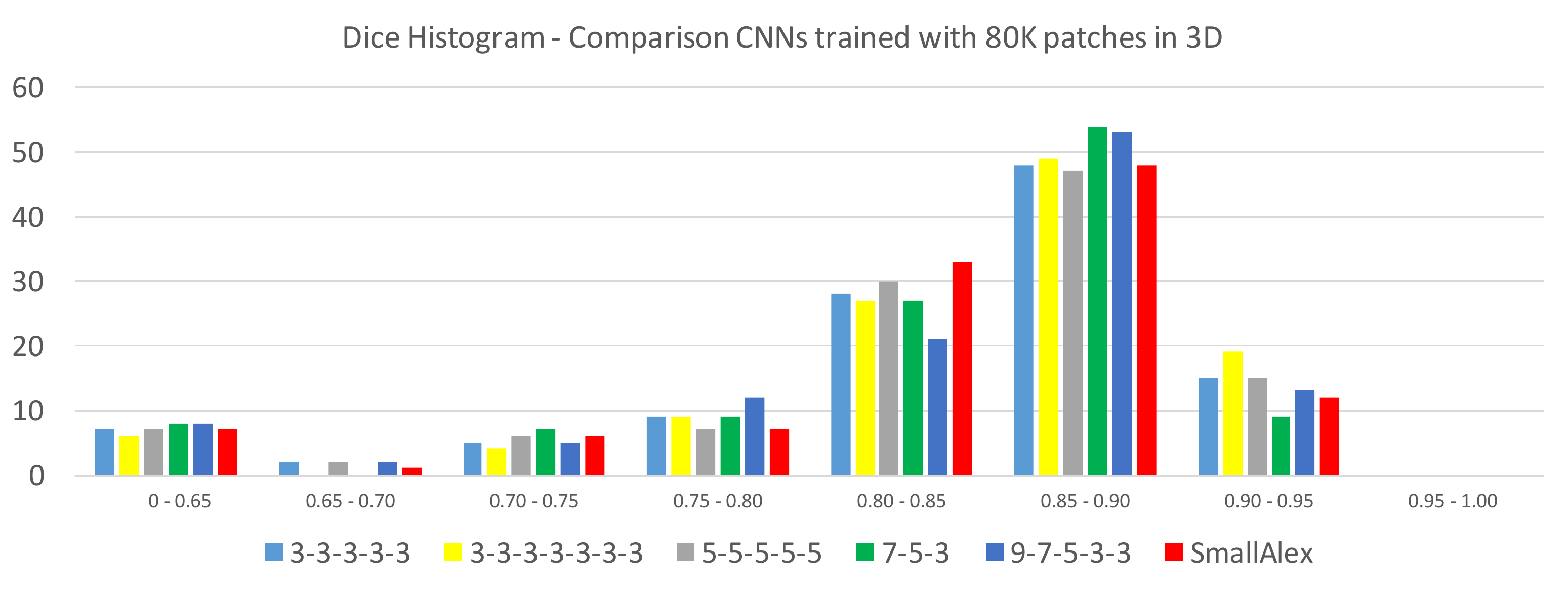} 
	\caption{The midbrain segmentation performance of each network on $114$ TCUS test volumes, under different training conditions, is summarised through histograms. The horizontal axis is subdivided in Dice bins having a width of $0.05$ Dice. The vertical axis represents the number of volumes falling in each Dice bin. Each CNN architecture is  depicted with its own colour.}
	\label{fig:US_diceHist}
\end{figure*}

Table \ref{table:USresults} shows the average performance in terms of Dice coefficients, mean distances of the estimated contours to the ground truth annotations and failure rates of the proposed Hough-CNN segmentation approach when different CNNs are employed. Since we segment one region per volume, the failure rate represents the percentage of volumes where the region of interest could not be segmented \rev{due to wrong localisation (Dice $0$)}. In Figure \ref{fig:US_diceHist} we provide summary of the performances of each network, when various amounts of training data are used and patches of different dimensionality are supplied. Better networks produce Dice histograms whose higher values are occurring far away from the origin.

Visual examples of ultrasound segmentation results are visible in Figure \ref{fig:SemanticVsHoughUS}. 
It is notable that the Hough-CNN segmentation is able to localise and segment the midbrain accurately, regardless of whether the scan was acquired through the left or right bone window. It is also robust to bone window quality and overall visibility of structures, as well as signal-drop regions and blurring.

\subsection{Experiments and results in MRI}
\label{sec:ResultsMRI}
We train each of our networks nine times (eight for 3D) in order to explore all the possible combination of different data dimensionality and size of the training set as explained in Section \ref{sec:datasets}. We test each of the models on $10$ volumes\rev{, using their respective atlas-based annotations for evaluation. We verified, through visual inspection performed by a technician and an expert neurologist, that the annotation appropriately delineate the regions of interest.}

Table \ref{table:mri_total} reports the average performance in terms of Dice coefficients, mean distances of the estimated contours to the ground truth annotations and failure rates of the proposed Hough-CNN segmentation approach when different CNNs are employed at its core. The failure rate, in particular, refers to the percentage of regions of the whole training set (total number: $26 \times 10$ regions), that were not segmented correctly by Hough-CNN \rev{due to the fact that they could not be correctly localised}. The results are clustered by the size of the training set employed to train the model to improve readability and the possibility of making comparisons between CNNs employing data having different dimensionality ($2$D, $2.5$D and $3$D). \rev{From these results we observe that the best performing architecture is ``7-5-3''}. 

\begin{sidewaysfigure}
	\centering
	\includegraphics[width=\linewidth]{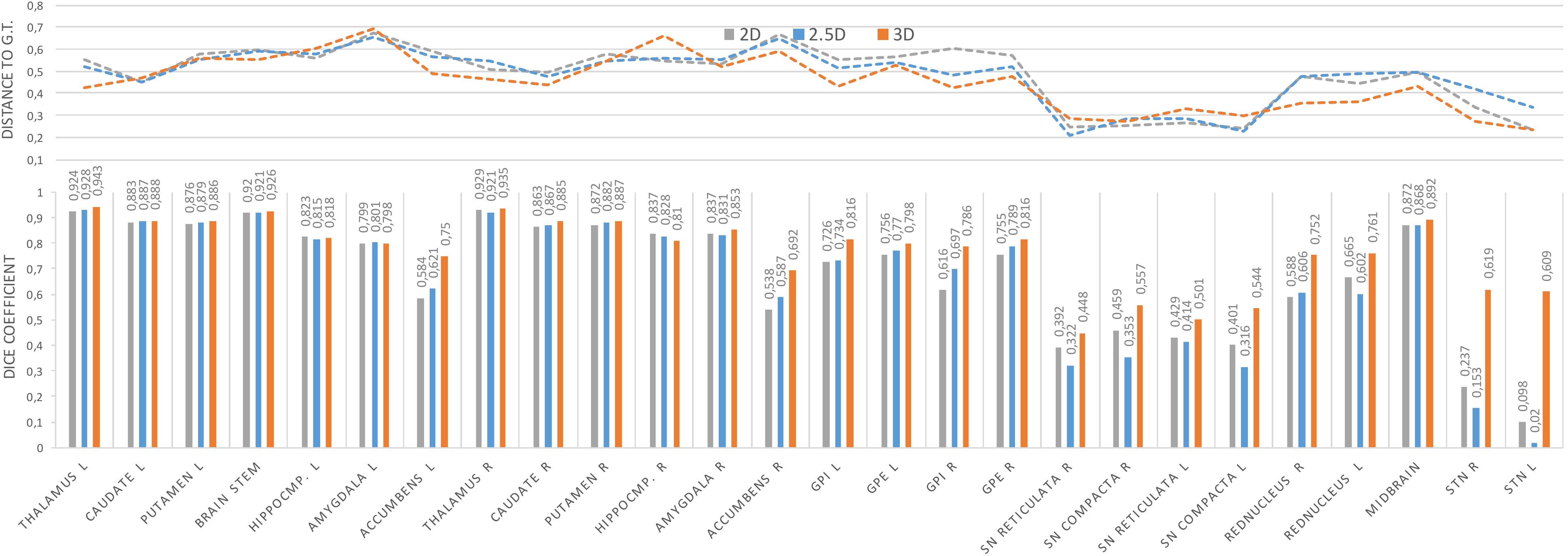}
	\caption{\rev{Average Dice coefficients (bar-plot) and distances to ground-truth delineation (dashed-lines plot), obtained segmenting the MRI test volumes using the best-performing network architecture ``7-5-3''. Dice coefficients are shown for each of the 26 target regions. Results obtained considering 2D, 2.5D and 3D data are represented in grey, blue and orange respectively. Best segmentation were delivered when 3D data was fed into the network, although the model was trained with only 1.35 millions 3D patches instead of the 13.5 million patches that were employed to train the models dealing with 2D and 2.5D data.}}
	\label{fig:MRI_biggest}
\end{sidewaysfigure}

\rev{In Figure \ref{fig:MRI_biggest} we compare the results achieved by the architecture ``7-5-3'', on each of the 26 brain region of interest separately, when different data dimensionalities are used. The bar plot shows the results in terms of Dice coefficient, while the dashed line plot conveys the results in terms of average distance of the estimated contour to ground-truth delineation. We observe that Hough-CNN yields better Dice coefficients when bigger regions and high contrast area are segmented. Small and low contrast regions could be correctly localised but they were in general harder to segment.}

Visual examples of MRI segmentation results are visible in Figure. \ref{fig:SemanticVsHoughMRI}. It is notable that the Hough-CNN segmentation is able to correctly localise and segment multiple structures, despite large anatomical variability, such as cortical atrophy and enlarged lateral ventricles. 

\begin{figure}
	\centering
	\includegraphics[scale=0.56]{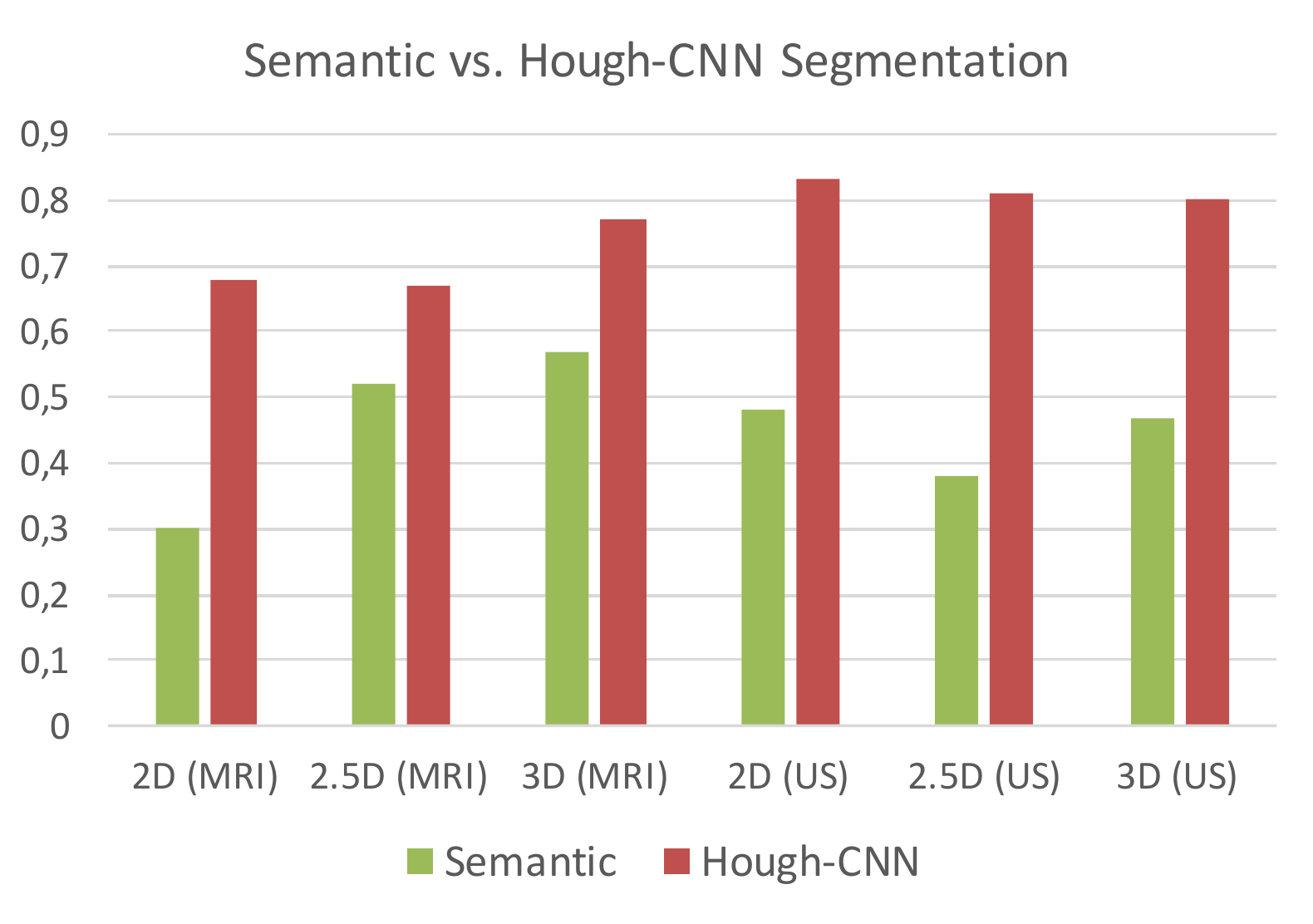}
	\caption{\rev{Comparison of mean Dice coefficients obtained in $2D$, $2.5D$ and $3D$ on US and MRI data using Hough-CNN and semantic segmentation.}}
	\label{fig:MRI_pixelwise}
\end{figure}

\begin{sidewaystable}
\begin{centering}
\begin{tabular}{|c|c|c|c|c|c|c|c|c|c|}
\hline 
Dimensionality $\rightarrow$ & \multicolumn{3}{c|}{2D} & \multicolumn{3}{c|}{2.5D} & \multicolumn{3}{c|}{3D}\tabularnewline
\hline 
Averages $\rightarrow$ & Dice $[0,1]$ & Distance (mm) & Failures & Dice $[0,1]$ & Distance (mm) & Failures & Dice $[0,1]$ & Distance (mm) & Failures\tabularnewline
\hline 
\hline 
\multicolumn{10}{|c|}{Training set size: $135$K patches}\tabularnewline
\hline 
3-3-3-3-3 & 0.61 & 0.52 & 6\% & 0.62 & 0.51 & 3\% & \textbf{0.70} & 0.46 & 0\%\tabularnewline
\hline 
3-3-3-3-3-3-3-3 & 0.61 & 0.52 & 8\% & 0.61 & 0.51 & 5\% & \textbf{0.70 }& 0.45 & 0\%\tabularnewline
\hline 
5-5-5-5-5 & 0.64 & 0.49 & 6\% & 0.63 & 0.52 & 1\% & \textbf{0.71} & 0.44 & 1\%\tabularnewline
\hline 
7-5-3 & \underline{0.67} & 0.48 & 4\% & \underline{0.68} & 0.48 & 2\% & \underline{\textbf{0.76}} & 0.45 & 0\%\tabularnewline
\hline 
9-7-5-3-3 & 0.60 & 0.52 & 8\% & 0.61 & 0.52 & 3\% & \textbf{0.68} & 0.49 & 0\%\tabularnewline
\hline 
SmallAlex & 0.61 & 0.53 & 5\% & 0.62 & 0.52 & 5\% & \textbf{0.71} & 0.46 & 0\%\tabularnewline
\hline 
\hline 
\multicolumn{10}{|c|}{Training set size: $1.35$M patches}\tabularnewline
\hline 
3-3-3-3-3 & 0.63 & 0.51 & 3\% & 0.62 & 0.52 & 5\% & \textbf{0.72} & 0.45 & 0\%\tabularnewline
\hline 
3-3-3-3-3-3-3-3 & 0.63 & 0.51 & 3\% & 0.62 & 0.52 & 2\% & \textbf{0.70} & 0.52 & 0\%\tabularnewline
\hline 
5-5-5-5-5 & 0.64 & 0.51 & 3\% & 0.61 & 0.52 & 3\% & \textbf{0.71} & 0.44 & 0\%\tabularnewline
\hline 
7-5-3 & \underline{0.68} & 0.47 & 2\% & \underline{0.68} & 0.47 & 2\% & \underline{\textbf{0.77}} & 0.45 & 0\%\tabularnewline
\hline 
9-7-5-3-3 & 0.63 & 0.53 & 4\% & 0.62 & 0.52 & 2\% & \textbf{0.68} & 0.47 & 1\%\tabularnewline
\hline 
SmallAlex & 0.64 & 0.51 & 4\% & 0.62 & 0.53 & 6\% & \textbf{0.72} & 0.46 & 0\%\tabularnewline
\hline 
\hline 
\multicolumn{10}{|c|}{Training set size: $13.5$M patches}\tabularnewline
\hline 
3-3-3-3-3 & \textbf{0.64} & 0.52 & 3\% & \textbf{0.64} & 0.52 & 3\% & \multirow{6}{*}{--} & \multirow{6}{*}{--} & \multirow{6}{*}{--}\tabularnewline
\cline{1-7} 
3-3-3-3-3-3-3-3 & \textbf{0.65} & 0.56 & 2\% & \textbf{0.65} & 0.54 & 0\% &  &  & \tabularnewline
\cline{1-7} 
5-5-5-5-5 & \textbf{0.64} & 0.51 & 2\% & \textbf{0.64} & 0.51 & 2\% &  &  & \tabularnewline
\cline{1-7} 
7-5-3 & \underline{\textbf{0.68}} & 0.49 & 3\% & \underline{0.67} & 0.48 & 3\% &  &  & \tabularnewline
\cline{1-7} 
9-7-5-3-3 & \textbf{0.63} & 0.52 & 5\% & \textbf{0.63} & 0.52 & 5\% &  &  & \tabularnewline
\cline{1-7} 
SmallAlex & \textbf{0.65} & 0.52 & 4\% & 0.63 & 0.53 & 5\% &  &  & \tabularnewline
\hline 
\end{tabular}

\par\end{centering}
\caption{Average segmentation results of 26 structures in 10 MRI test volumes, using Hough-CNN with variations of architectures (single rows), patch dimensionalities (column blocks) and training set sizes (row blocks). \rev{The best result for each architecture (across the data dimensionalities) are highlighted by using bold typeface. The best results for each dimensionality (across the architectures) are underlined. The best result is obtained using the architecture ``7-5-3'' and 3D data.}}
\label{table:mri_total}
\end{sidewaystable}

\section{Discussion}
\label{sec:Discussion}

Training of CNNs requires a large amount of data in order to achieve satisfactory voxel-wise classification results and perform semantic segmentation. However, as described in the introduction, obtaining such large annotated datasets is rarely possible in clinical settings. By using a voting-based strategy, it is possible to localise the anatomy of interest with high precision, even when the rate of mis-classified voxels is very high. Additionally, our Hough-CNN approach implicitly enforces shape priors which facilitate segmentations in images where the anatomy of interest is poorly visible. Furthermore, when using 3D patches, only 1.35M training patches were required to surpass the performance obtained with datasets of 13.5 millions 2D and 2.5D patches. This marks a 90\% reduction of required training data. In all three dimensionalities, 2D, 2.5D and 3D, Hough-CNN outperforms voxel-wise segmentation (cf. Figure \ref{fig:MRI_pixelwise}). Similar to related works \cite{ngo13, ranftl14, turaga10, liang15}, we thus demonstrate that it may be beneficial to embed CNNs as powerful classifiers into higher-level methods which encode anatomic shape- and appearance priors. 

The experiments performed on MRI highlight important aspects of both our CNNs and the modality itself. Most of the brain regions considered in this study (e.g. midbrain, STN, caudate) can be recognised by a human rater by clearly visible contrasts, while \rev{the position and boundaries of difficult regions with less contrast (e.g. GPi, GPe, SNpc, SNpr)} can be inferred \rev{through anatomical knowledge and neighborhood context}. Ultrasound volumes are much more challenging from this point of view. Human midbrain in TCUS \rev{can be difficult to discern} and human observers can be mislead by artefacts and signal-loss areas having similar shape. The CNNs employed in this study had various architectures and therefore different pattern recognition capabilities. In MRI, where \rev{the most part of regions of interest have good contrast while the position of the others can be inferred by the context}, the best performing network was ``7-5-3''. Although this architecture is the simplest, it delivered best results in all the MRI experiments. In US, which is a challenging modality, the networks that delivered best results were among the most complex. ``SmallAlex'' and ``3-3-3-3-3-3-3-3'' are deeper and therefore recognise more complex visual content than ``7-5-3''.

While we observed a strong performance advantage when segmenting MRI volumes considering 3D data (Table \ref{table:mri_total}), we observed the opposite effect when segmenting ultrasound as shown in the bottom left of Table \ref{table:USresults}. In MRI, processing data in 3D brings additional useful information which improves the performance of both automated methods and human raters, who refer simultaneously to sagittal, coronal and axial views when establishing the ground truth. In US, we observed that experts segmenting the ground truth used only the axial plane, since it is the only plane in which the characteristic shape of the midbrain can be recognised. Similarly, CNNs produce best results when they are not supplied with misleading information from sagittal and coronal planes. 

Altogether, using Hough-CNNs, we segmented 10 previously unseen MRI volumes achieving very high Dice coefficients \rev{for large and high-contrasted} regions, while \rev{some of the smallest and most challenging regions were almost always} localised accurately \rev{and segmented with sub-voxel mean surface distance.} Additionally, we
achieved very robust midbrain segmentation in 3D-TCUS, in a test dataset of more than 20 subjects and 114 volumes, with a large variation of \rev{3D sweep geometry}, bone window qualities, midbrain appearance, location and orientation. Given the size and variety of the 3D-TCUS test set, we are confident to say that the method generalises well to unseen patients.

\rev{Compared to atlas-segmentation, Hough-CNN is faster (30 seconds in US, and 3-4 minutes in MRI on the machine employed for testing) and entirely registration-free. This makes our approach applicable to TCUS data, in which registration-dependent methods like atlas-based segmentation would be extremely difficult, if not impossible, due to largely missing anatomical and structural context. Our approach is flexible since both votes and segmentation patches can be substituted without any need for re-training or augmented to include information from multiple experts. As a future work, we plan to investigate the extendability of the trained CNN classifier to other modalities via transfer learning, e.g. from our QSM sequences to T1 or T2.  It is also noteworthy that in this work, we have only used the CNN method for segmentation. However, as other works have demonstrated \cite{payan15}, the learned data representations in the last layers of the CNN can be directly used for classification or regression of disease parameters. This can be interleaved with segmentation, which goes far beyond the capabilities of purely atlas-based methods.}

\section{Conclusion}
In this work, we applied CNNs to medical image segmentation, under the constraints of limited training data and computational resources. We performed a large study of several CNN parameters, including architectures, patch dimensionality and training set size, highlighting CNN performance given challenges from different modalities. We proposed Hough-CNN, a patch-wise multi-atlas method which implicitly encodes priors on anatomic shape and context. The method outperformed voxel-wise semantic segmentation of CNNs in all parameter settings, while using less training data and delivering smooth segmentation contours without the need for post-processing. The method is modality-independent and scalable to multiple regions and harnesses the impressive classification power of CNNs and Deep Learning for application in clinical settings.


\section*{Acknowledgment}
 This study was funded by the L\"{u}neburg Heritage and  Deutsche Forschungsgesellschaft (DFG) Grant BO 1895/4-1. We gratefully acknowledge the support of NVIDIA Corporation in donating a ``Tesla K40'' GPU for this study.

\bibliographystyle{elsarticle-num}
\bibliography{elsa-cnn-bibliography}

\end{document}